\pdfoutput=1

\documentclass[11pt]{article}

\usepackage[]{ACL2023}

\usepackage{times}
\usepackage{latexsym}

\usepackage[T1]{fontenc}

\usepackage[utf8]{inputenc}

\usepackage{microtype}

\usepackage{url,booktabs}
\usepackage{array,multirow,graphicx}
\usepackage{subcaption,amsmath}
\usepackage{xspace,mfirstuc,tabulary}

\newcommand{\ar}{\texttt{Resolution}\xspace}
\newcommand{\hi}{\texttt{Human Interest}\xspace}
\newcommand{\co}{\texttt{Conflict}\xspace}
\newcommand{\ec}{\texttt{Economic}\xspace}
\newcommand{\mo}{\texttt{Moral}\xspace}

\newcommand{\vil}{\texttt{Villain}\xspace}
\newcommand{\her}{\texttt{Hero}\xspace}
\newcommand{\vic}{\texttt{Victim}\xspace}

\title{Conflicts, Villains, Resolutions:\\ Towards models of Narrative Media Framing}

\author{Lea Frermann$^1$ \hspace{3ex} Jiatong Li$^2$ \hspace{3ex} Shima Khanehzar$^1$$^{,3}$ \hspace{3ex} Gosia Mikolajczak$^4$ \\
        $^1$School of Computing and Information Systems, The University of Melbourne \\ 
        $^2$Department of Computing, The Hong Kong Polytechnic University 
        $^3$CSIRO Data61 \\
        $^4$The Global Institute for Women's Leadership, The Australian National University\\
        \begin{tabular}{p{0.5\linewidth}lp{0.5\linewidth}}
        \multicolumn{1}{c}{\texttt{lfrermann@unimelb.edu.au}} && \multicolumn{1}{c}{\texttt{jiatong.li@connect.polyu.hk}}\\
        \multicolumn{1}{c}{\texttt{shima.khanehzar@data61.csiro.au}} && \multicolumn{1}{c}{\texttt{Gosia.Mikolajczak@anu.edu.au}}
        \end{tabular}
        }

\begin{document}

\maketitle
\begin{abstract}
Despite increasing interest in the automatic detection of media frames in NLP, the problem is typically simplified as single-label classification and adopts a topic-like view on frames, evading modelling the broader document-level narrative. In this work, we revisit a widely used conceptualization of framing from the communication sciences which explicitly captures elements of narratives, including {\it conflict} and its {\it resolution}, and integrate it with the narrative framing of key entities in the story as {\it heroes}, {\it victims} or {\it villains}. We adapt an effective annotation paradigm that breaks a complex annotation task into a series of simpler binary questions, and present an annotated data set of English news articles, and a case study on the framing of climate change in articles from news outlets across the political spectrum. Finally, we explore automatic multi-label  prediction of our frames with supervised and semi-supervised approaches, and present a novel retrieval-based method which is both effective and transparent in its predictions. We conclude with a discussion of opportunities and challenges for future work on document-level models of narrative framing.\footnote{Code and data are available at \url{https://github.com/phenixace/narrative-framing}.}
\end{abstract}

\section{Introduction}
Media discourse around contested issues is often biased by experiences or interests of the news outlets and different stakeholders they give voice to. News framing by the media has been formalized and examined on many levels in communication and social sciences, ranging from selection of information~\cite{levitt1981whole} over discourse-centric~\cite{pan1993framing} and entity focused approaches~\cite{lawlor2017deciding}. While a growing body of work in NLP attempts to automatically detect framing in the news or social media, most work adopts well-defined yet oversimplifying approaches like topic modeling or simple classifiers~(see~\citet{ali-etal-2022-survey} for a recent review); formalize the task as single-label classification ignoring co-existence and interactions of different frames, and focus on localized emphasis frames rather than the full story~\cite{card-etal-2015-media,field-etal-2018-framing,khanehzar2021framing}. 

\begin{table}[t]
    \centering
    \begin{small}
    \setlength{\tabcolsep}{1pt} 
    \begin{tabular}{p{1\columnwidth}}
    \toprule
        \ar: solution/alleviation of the issue  \\
        \co: disagreements between individuals, groups, institutions, countries, etc.  \\
        \hi: emotionalization and dramatization of an issue through the lens of affected individuals  \\
        \mo: moral or religious references \\
        \ec: economic consequences for individuals, groups, institutions, countries, etc. \\\midrule
        \her: entity contributing to/responsible for issue resolution \\
        \vil: entity contributing to/responsible for issue cause\\
        \vic: entity suffering the consequences of an issue\\\bottomrule
    \end{tabular}
    \end{small}
    \caption{Five frames (top) and three narrative roles (bottom) considered in this paper.}
    \label{tab:frames}
\end{table}

This paper addresses the above shortcomings by considering framing through the lens of narratives. We adopt a small set of high-level framing devices established in the communication literature~\cite{neuman1992common,semetko2000framing}, and integrate them with narrative roles assigned to key actors (entities) in the discourse. Table~\ref{tab:frames} defines the frames and associated entity roles. We argue that more nuanced and transparent automated models of framing are essential to meaningfully support social studies into a systematic understanding of the viewpoints presented by different stakeholders in contested issues such as climate change. Our contributions to this end are (1)~introducing an established frame inventory and annotation procedure from the communication sciences into NLP; (2)~a labeled data set; (3)~a case study on the framing of climate change, showcasing the potential of our annotations for large-scale media analysis; and (4)~experiments on automatic frame prediction, including an effective and transparent retrieval-based classifier to predict multiple frames per article.

Table~\ref{tab:frames} (top) summarizes the frames used in our work. Our framework departs from existing NLP approaches in two ways: first, we adopt a multi-label classification paradigm, allowing for a more nuanced analysis and avoiding the oversimplification of framing to a single label per item. Secondly, our framework emphasizes the narrative structure of the {\it full} article: frames such as {\it conflict}, {\it resolution} or {\it human interest} are central building blocks of narratives, and are dominant in news coverage of contested issues~\cite{semetko2000framing}. Building on components of the Narrative Policy Framework~\cite{shanahan2018narrative}, we identify the key entities responsible for the issue ({\it villains}), those who are affected ({\it victims}) and those who can resolve the issue ({\it hero}); see Table~\ref{tab:frames} (bottom). 

We apply our framework to the issue of climate change, a pressing global challenge with wide-reaching impacts~\cite{pewresearchcenter_2022}, which remains politically contested in terms of the understanding of its urgency, causes, and possible solutions~\cite{sparkman2022americans}. Importantly, studies show that climate `skeptics' (and those lacking scientific backing) are cited almost twice as often in the mainstream news media as those calling for climate action~\cite{wetts2020climate}, rendering the examination of media framing and its effects on public support for climate change mitigation a pressing goal. 
While a substantive body of work on climate change framing emerged in the social and communication sciences~\cite{nisbet2009communicating,shanahan2022narrative}, the issue has attracted surprisingly little attention in the NLP community to date. Exceptions include work on stance detection in news~\cite{luo2020detecting} or social media~\cite{vaid-etal-2022-towards} or models of scepticism detection~\cite{bhatia-etal-2021-automatic}, whereas we focus on the narrative framing of the issue across political leanings. To recap, in this paper we present:
\begin{itemize}
 \item The concept of ``Narrative Media Framing'' formalized through a set of frames about conflicts, their effects and resolutions which are integrated with narrative roles assigned to key actors~(Section~\ref{sec:annotation}).
 \item The narrative frames corpus of 428 English news articles on climate change labeled with frame devices (Table~\ref{tab:frames}). Following~\citet{semetko2000framing}, annotators answered binary indicator questions, and the final frame labels were derived from the answer set~(Section~\ref{sec:annotation}). 
 \item A detailed analysis of our annotated data set, highlighting the interaction of frames and narrative roles, and differences across media outlets with different political bias (Section~\ref{sec:exploratory}). 
 \item Experiments on automatic frame prediction, including semi-supervised and supervised methods, including a new simple and transparent, yet effective method which combines retrieval with classification~(Section~\ref{sec:prediction}). 

\end{itemize}


\section{Background}
\label{sec:background}
Media framing refers to the deliberate presentation of information in order to elicit a desired response or shift in reader's attitude. We introduce into NLP five high-level frames (Table~\ref{tab:frames} top), identified by~\citet{semetko2000framing} as covering the dominant framing in reporting on contested issues with the aim to attract reader's attention~\cite{mendelsohn-etal-2021-modeling}. These categories have been applied via manual content analysis to a variety of issues and events, ranging from the media coverage of the Egyptian revolution~\cite{fornaciari2012framing}, over the MH370 crash~\cite{bier2018framing} to climate change~\cite{dotson2012media,feldman2017polarizing}. To identify each frame,~\citet{semetko2000framing} proposed a set of binary indicator questions which improved annotation quality and portability of the framework across studies. We construct the first publicly available data set annotated with this framework, cover a larger and more diverse set of news articles than prior work, and link the frames with narrative roles assigned to key entities appearing in the story. 

Media framing may manifest through the narrative roles -- \her, \vil or \vic\ -- assigned to key entities in a document, and this phenomenon has been widely studied in the communication sciences in general~\cite{shanahan2018narrative} and in the context of climate change in particular~\cite{luck2018counterbalancing}. We draw on this work as well as work which identified key stakeholder categories in the climate change discourse~\cite{haigh2009natural,ahchong2012anthropogenic,chen2022climate} to analyze the framing of entities in news articles along the political spectrum.

NLP studies on framing have predominantly focused {\it emphasis framing}, the strategic inclusion or omission of aspects of an issue, such as legality or public opinion~\cite{card-etal-2015-media} or, to a lesser extent, on {\it equivalence framing} as different expressions of identical concepts (``alien'' vs ``immigrant'', ~\citet{lee-etal-2022-neus,ziems-etal-2022-inducing}). Both perspectives focus on local, lexical signals. Emphasis framing is typically formalized as a single-label prediction of the most dominant frame in a news article~\cite{card-etal-2015-media}, headline~\cite{liu-etal-2019-detecting,akyurek-etal-2020-multi}, or a social media post~\cite{johnson2017modeling,hartmann-etal-2019-issue}. The media frames corpus~\cite{card-etal-2015-media} is one of the most comprehensive frame-labeled data sets comprising several thousand news articles across five  contested issues. While the data includes span-level labels which {\it could} be used for multi-label classification, work using the MFC predominantly attempts document-level prediction of a single ``primary'' article frame, disregarding span labels~\cite{ji-smith-2017-neural,khanehzar2021framing}.\footnote{Although see~\citet{field-etal-2018-framing} for an exception.} More broadly, our work complements emphasis frames by considering more abstract frames around conflict, resolution, and personal, moral and economic impacts. Both formalizations of framing have a strong foundation in the communication literature, and studying their interaction at scale with NLP methodology is an interesting avenue for future work.

\citet{mendelsohn-etal-2021-modeling}~consider a variety of framing strategies in the context of tweets, but with less of a focus on story structure due to the short document lengths. Our framework complements their work in three ways: i)~we approach framing as multi-label classification, relaxing the assumption of a single frame per article; ii)~we present a set of frames that are abstract with evidence distributed across a document, requiring higher-level document model; and iii) we link frames with narratives via entity roles in a unified annotation framework consisting of a series of binary indicator questions, allowing us to study the interplay of framing and narratives.

\begin{table*}[]
    \centering
    \begin{small}
    \begin{tabular}{lp{0.9\linewidth}}
    \toprule
         \multirow{2}{*}{\rotatebox{90}{RE}} & (1) Does the story suggest a solution(s) to the issue/problem?; (2)  Does the story suggest that some entity could alleviate the problem? \\\midrule
         \multirow{3}{*}{\rotatebox{90}{CO}} & (1) Does the story reflect disagreement between political parties/individuals/groups/countries?; (2) Does one party/individual/group/country reproach another?; (3)  Does the story refer to two sides or more than two sides of the problem or issue? \\\midrule
    \multirow{3}{*}{\rotatebox{90}{HI}} & (1) Does the story provide a human example or a "human face" on the problem/issue?; (2) Does the story employ adjectives or personal vignettes that generate feelings of outrage, empathy-caring, sympathy, or compassion?; (3) Does the story go into the private or personal lives of the entities involved?\\\midrule
    \multirow{2}{*}{\rotatebox{90}{MO}} & (1) Does the story contain any moral message?; (2) Does the story make reference to morality, God, and other religious tenets?\\\midrule
    \multirow{2}{*}{\rotatebox{90}{EC}} & (1) Is there a mention of financial losses or gains now or in the future?; (2) Is there a mention of the costs/degree of the expense involved?; (3) Is there a reference to the economic consequences of pursuing (or not) a course of action? \\\bottomrule
    \end{tabular}
    \end{small}
    \caption{Binary indicators for the five frames: \ar (RE), \co (CO), \hi (HI), \mo (MO), \ec (EC).}
    \label{tab:questions}
\end{table*}


\section{The Narrative Frames Corpus}
\label{sec:annotation}
We identified 17.9K English-language news articles on climate change published in 2017--2019 in the UK and in the US by matching a set of climate change-specific keywords in articles from the NELA corpora~\cite{nela2017,nela2018,nela2019}. {See Appendix~\ref{app:keywords} for more details.} For each article, NELA provides metadata about its publication date, media outlet, and its associated political leaning as identified by the Media Bias Fact Check (MBFC) website.\footnote{\url{https://mediabiasfactcheck.com/}} We manually annotated a subset of 428 articles of this data set, balanced across the three years and the four most dominant MBFC categories: left, center-left, right and questionable source.\footnote{The `center-right' category was very rare in the set of sampled articles, and hence merged with `right'.}

We recruited four on-site annotators, all English native speakers with a background in the social/political sciences. The annotators went through an extensive training phase including several rounds of feedback. Details of annotator remuneration can be found in the Ethics statement. 

\paragraph{Frame annotations} We adapted \citet{semetko2000framing}'s frame indicator questions. We added a pre-screening question to confirm that an article is predominantly ($>$70\%) about climate change, removed one question about visual information (as we focus on text only), and changed wording specific to the `government' to `any entity' to align with our broad definition of stakeholder entities, discussed below. The full questionnaire is shown in Appendix~\ref{app:codebook}. Annotators were presented with the full article text together with the questionnaire, but no explicit meta information such as outlet name or date of publication.

The raw annotations provided answers to a list of binary indicator questions. We verified that the mapping (i.e., the factor structure) between the five frames in Table~\ref{tab:frames} and their associated indicator questions in~\citet{semetko2000framing} replicates in our annotated data set. To do so we ran a confirmatory factor analysis (CFA;~\citet{brown2012confirmatory}).\footnote{The input data for the CFA is based on majority voting, i.e., at least two out of three annotators agreeing on a given response. Prior to the main analysis, we converted raw dichotomous (0-1) scores into a polychoric correlation matrix which served as an input into CFA.} We removed all items with a factor loading $<0.3$ (as not fitting well into any of the five factors), retaining a total of 13 indicators, with 2-3 indicators loading on a given frame. These questions are listed in Table~\ref{tab:questions}. The final model fitted the data well (CFI=.945; RMSEA=.052[.039, .065], p=.370; SRMR=.059), confirming the five-factor structure.
An article was then labeled with a frame if $\geq 2$ indicator questions for that frame were answered `yes' by $\geq2$ annotators.\footnote{Except for \mo, where only 1 question had be answered `yes' by a majority of annotators due to the rarity of the label.} This resulted in a multi-label data set with articles covering zero (12\%), one (39\%), two (32\%) three (15\%) or four (3\%) frames. See  Appendix~\ref{app:labelstats} for additional data set statistics.

\paragraph{Entity annotations} Entities were annotated as part of the binary indicator questionnaire introduced above.  Three indicator questions assessed whether an article contained an entity that could {\it alleviate the problem} (\her); {\it was responsible for the problem} (\vil) or {\it was negatively affected by the issue} (\vic). If an annotator answered `yes' to any of these questions, they were asked to identify the most appropriate entity in the text. An entity was meant to be selected only if the article was explicit about that entity's role (e.g., a politician was depicted as 
"the only person who could save the planet") and strictly based on the entity's presentation in the article, rather than the annotator's opinion about that entity.\footnote{E.g., if an article presented Trump as a person who could mitigate climate change, the annotator was supposed to tag him as a \her, even if they didn't personally agree with that interpretation.} We included all entities extracted by our annotators as part of our published data set.

\paragraph{Annotator agreement} Krippendorff's $\alpha$ across four annotators and 13 frame indicator questions is 0.52, indicating fair agreement as expected for a complex task like frame annotation. Average pairwise agreement without chance-correction is 0.78 (min=0.75, max=0.81). 

A total of 2,185 entities were extracted across all narrative roles. Average pairwise agreement on {\it existence} of a role in an article was 0.59 (Krippendorff's $\alpha=0.40$). To assess agreement on the {\it identity} of entities for roles which were attested by at least two annotators, we computed the exact string match of associated entities, after basic text normalization. Entities match exactly 41\% of the time. We also computed more lenient metrics based on token overlap (average Rouge-L=0.45) and embedding similarity (average. BertScore=0.91) between pairs of extracted entities.

The agreement for both role detection and entity-role assignment was low overall, suggesting that the task is challenging. In this paper, we use the narrative role labels in the exploratory analysis in Section~\ref{sec:exploratory} and discuss future work on computational modeling of narrative roles in Section~\ref{sec:discussion}.

\subsection{Stakeholder categories} 
\label{ssec:stakeholders}
\begin{figure*}[ht]
\centering
\begin{subfigure}{\textwidth}
\includegraphics[width=\textwidth]{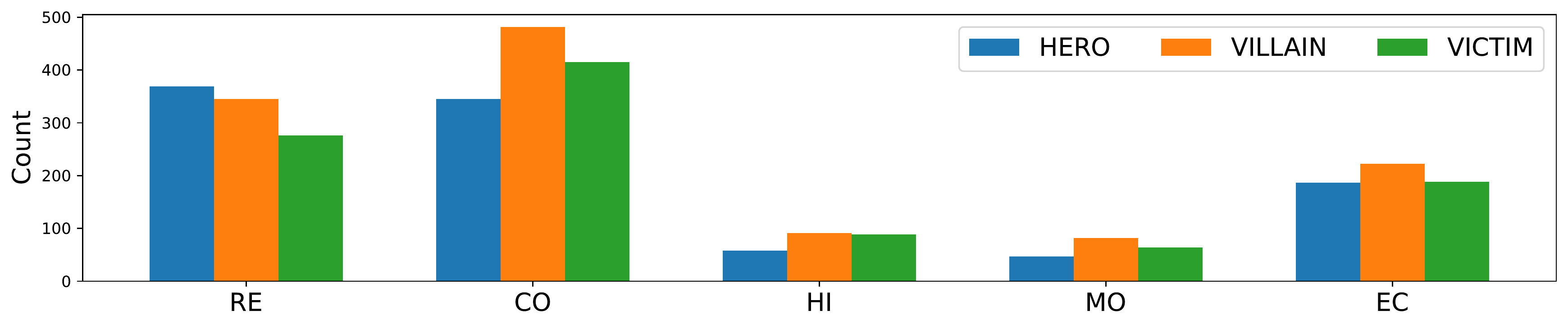}
    \caption{Co-occurrence (count) of roles with frames.}
    \label{sfig:roles:frame}
    \end{subfigure}
    \begin{subfigure}{\textwidth}
    \includegraphics[width=\textwidth]{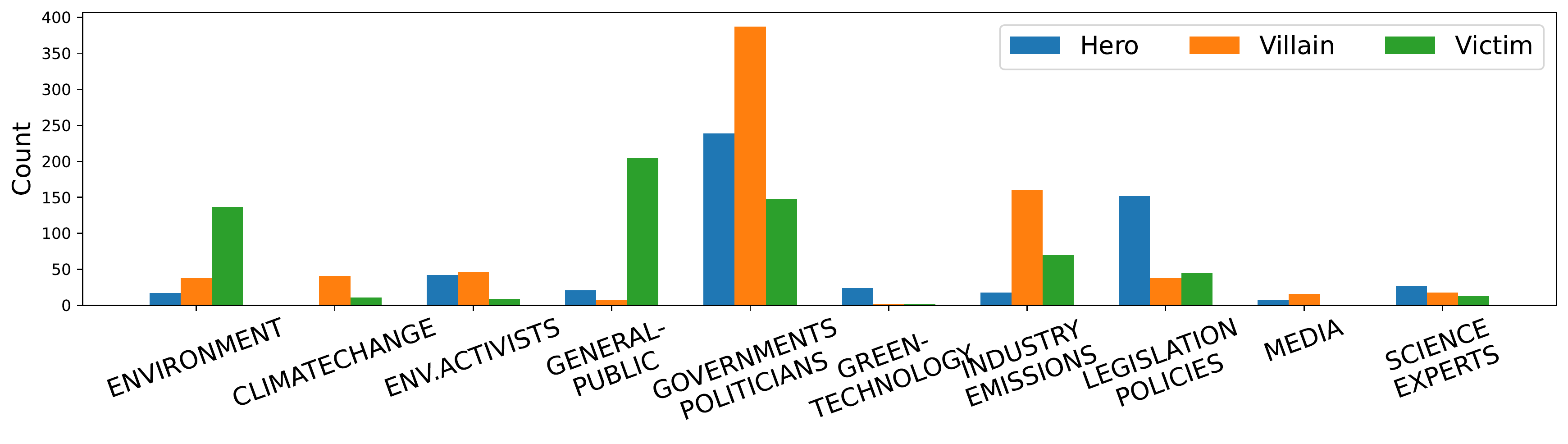}
    \caption{Co-occurrence (count) of roles with stakeholder groups.}
    \label{sfig:groups:rolesgroups}
    \end{subfigure}
    \caption{Association of roles with different frames (top) and stakeholder groups (bottom).}
    \label{fig:rolesframes}
\end{figure*}

We grouped the ${>}2K$ extracted entities into a smaller set of stakeholder categories to ease analysis. We identified 10 such categories from the previous literature~\cite{ahchong2012anthropogenic,blair2016applying,chen2022climate,haigh2009natural}, adopting a broad definition of stakeholders which includes groups or entities that `affect or are affected by' the issue of climate change~\cite{freeman1984stakeholder}. The set of stakeholder categories is shown in Figure~\ref{sfig:groups:rolesgroups}, and Appendix~\ref{app:entitygroups} provides additional details. One annotator assigned each unique extracted entity to its most appropriate stakeholder category (or a generic category `Other' if no other category fit). \\
\\
In sum, the narrative frames corpus consists of 428 English news articles labeled with (1)~multi-label frame categories; (2)~narrative roles for specific entities; and (3)~their associated stakeholder category; as well as meta-data about the article's date of origin, outlet, and associated political leaning. 

\begin{figure}[h]
    \centering
        \includegraphics[width=1.1\columnwidth]{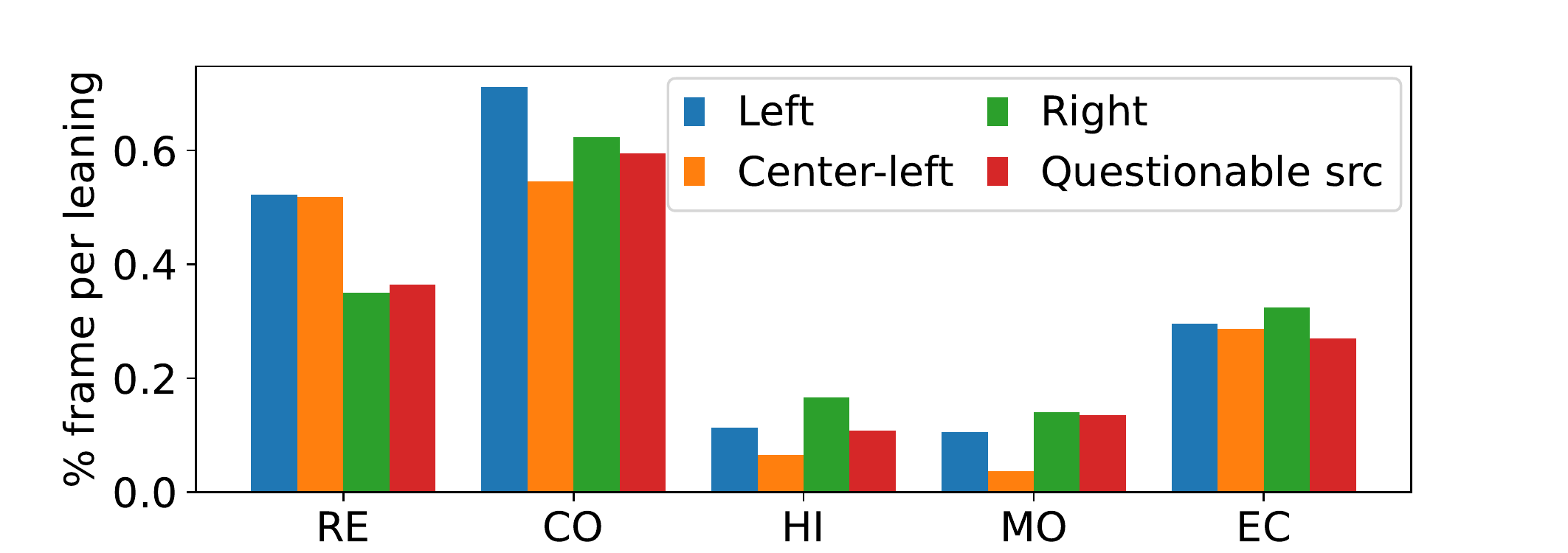} 
    \caption{Distribution of frames across political leanings of news outlets (as attested by MBFC).}
    \label{fig:exploratory}
\end{figure}

\section{Narrative Framing of Climate Change}
\label{sec:exploratory}


We conduct an exploratory analysis on the framing of climate change in media outlets with different political leaning, as well as the interplay of frames, narrative roles and stakeholder categories.

\paragraph{Framing and political leaning} Figure~\ref{fig:exploratory} shows the proportion of articles mentioning each frame by the media outlets' political leaning.\footnote{Noting that the numbers do not sum to one due to the multi-label nature of our annnotations.} 
\co (CO) and \ar (RE) are most prevalent across all leanings. The \mo frame (MO) is least prevalent throughout. This pattern is partially consistent with previous research. \citet{dirikx2010frame} found \ar, but not \co, to dominate in climate change reporting in the Netherlands and France in early 2000s, which might suggest that the discourse on climate change has become more polarized over time, in particular in our data set of US and UK news coverage where the media landscape is strongly partisan. For example, in a more recent study involving four major US newspapers, \citet{kim2018news} show that \co is the most common frame in the context of US immigration.

\ar (RE) is more prevalent in the left-leaning outlets (left, left\_center), while the opposite is true for \hi (HI): right-leaning (and questionable) outlets are more likely to refer to personal stories and use language evoking empathy. These findings are partially consistent with prior work, e.g,~\citet{feldman2017polarizing} show that {\it both} \ec and \co are more likely to be used in conservative outlets, while we find \co prevalent across the board. However, \citet{feldman2017polarizing} only included three major US news papers, in contrast to 41 in our analysis.


\begin{figure*}
    \begin{subfigure}{.33\textwidth}
    \includegraphics[width=\textwidth,clip,trim=0cm 0.3cm 0 0]{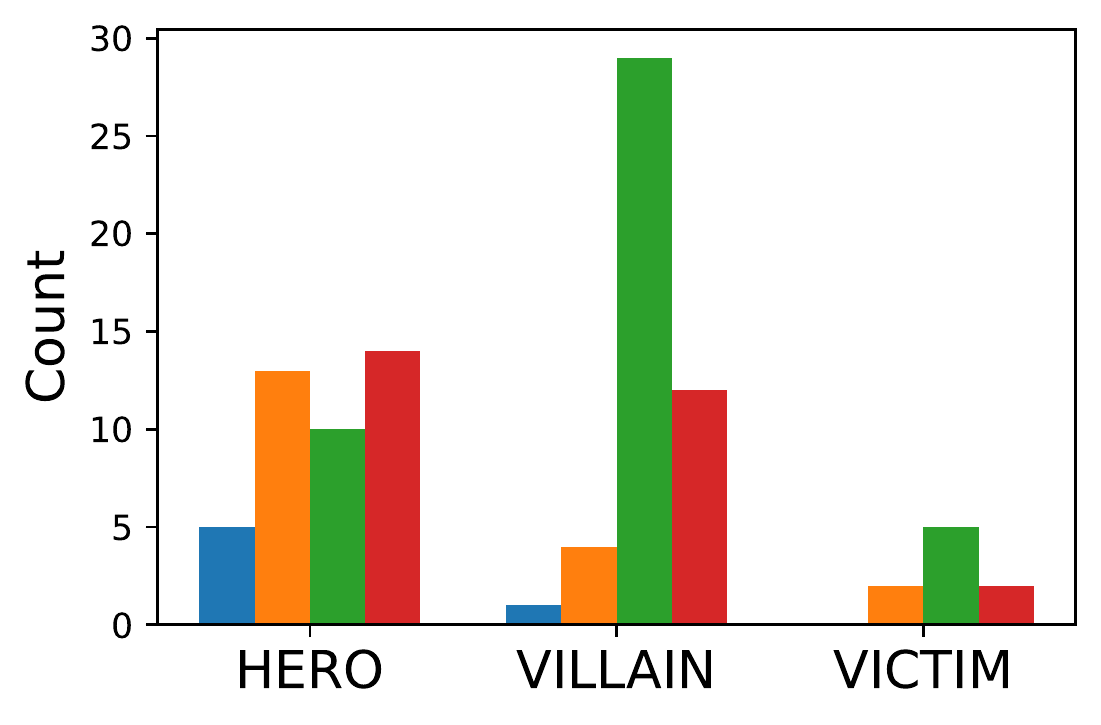}
    \caption{Environmental activists \& orgs.}
    \label{sfig:roles:envorg}
    \end{subfigure}
\begin{subfigure}{.33\textwidth}
    \includegraphics[width=\textwidth,clip,trim=0cm 0.3cm 0 0]{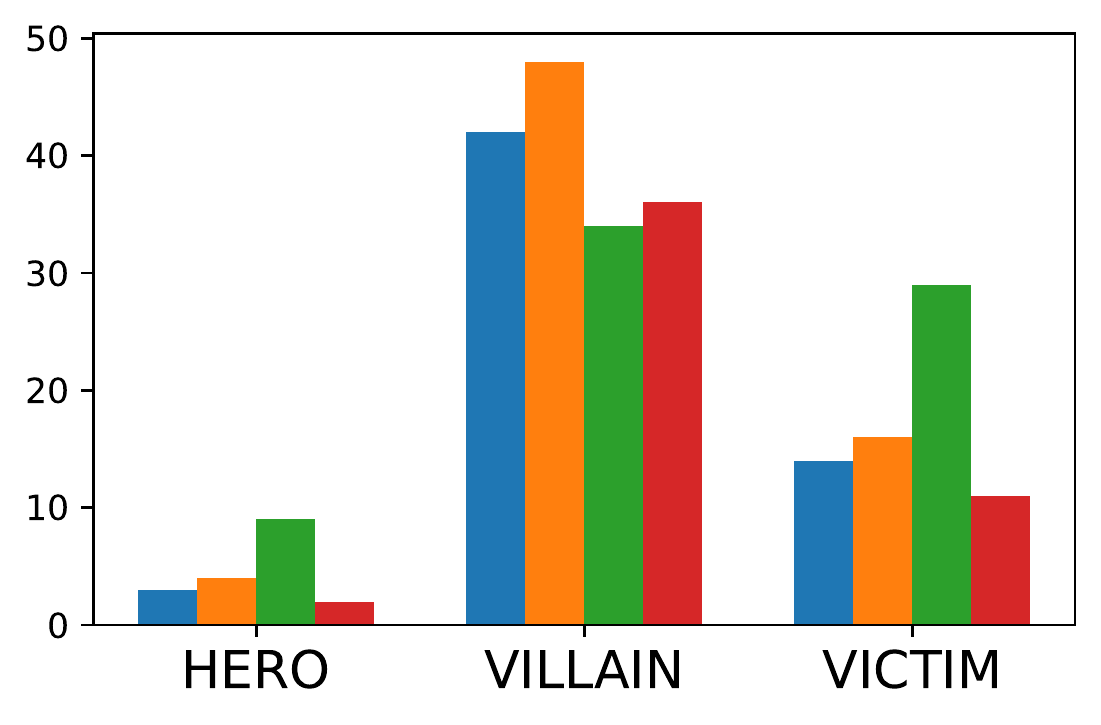} 
    \caption{Industry \& Emissions}
    \label{sfig:roles:industry}
    \end{subfigure}
        \begin{subfigure}{.33\textwidth}
    \includegraphics[width=\textwidth,clip,trim=0cm 0.3cm 0 0]{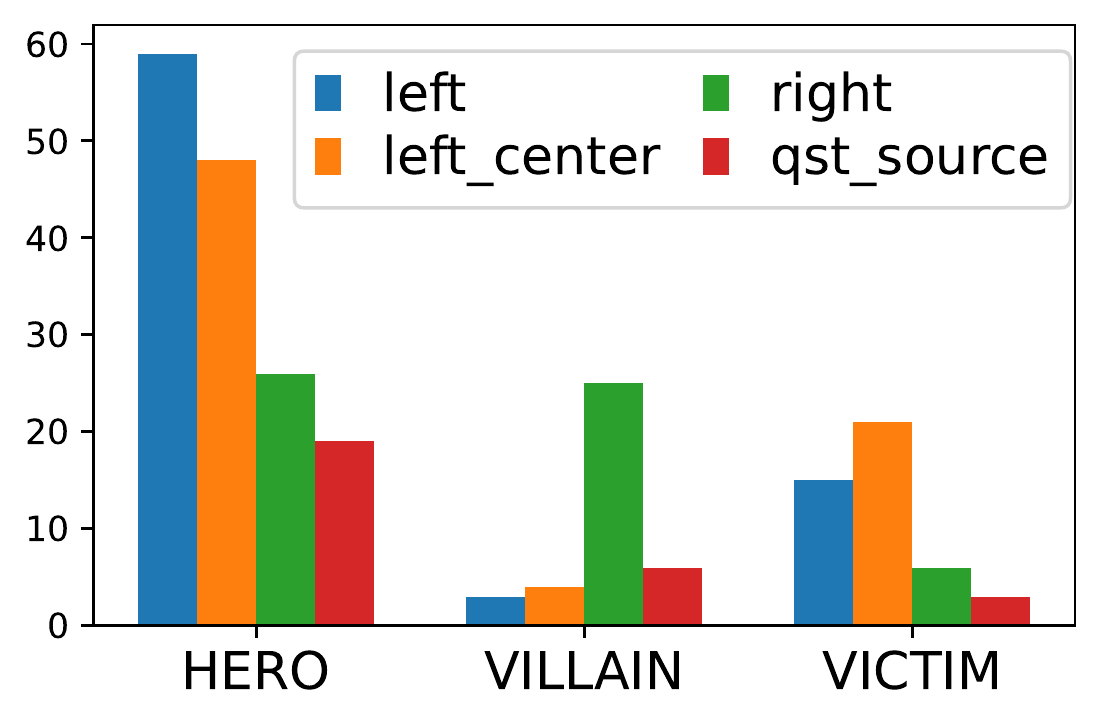} 
    \caption{Legislation \& Policies}
    \label{sfig:roles:legislation}
    \end{subfigure}
    \caption{Narrative roles assigned to stakeholder categories in news outlets with different political leaning.}
    \label{fig:exploratory:roles}
\end{figure*}

\paragraph{Frames, roles and stakeholders}
Figure~\ref{fig:rolesframes} illustrates the association of narrative roles with different frames (\ref{sfig:roles:frame}) and stakeholders (\ref{sfig:groups:rolesgroups}). Unsurprisingly, the \her, an entity presented with the ability to fix or alleviate the issue under discussion, is the most prevalent role in the \ar frame. The \vil dominates most other frames, except for the \hi frame where \vic is equally dominant. This aligns with a well-known ``negativity bias'' in news reporting, i.e., a dominance of negative content with a focus on problems, conflicts and their causes and victims~\cite{soroka2019cross}.

We explore the distribution of roles across stakeholder categories in Figure~\ref{sfig:groups:rolesgroups}. Overall, Governments \& Politicians are the most dominant stakeholder category, typically depicted as the \vil (of all stakeholder categories they are also most likely to be depicted as the \her pointing to ambivalent attitudes toward this category). The Environment and the General Public dominate the \vic role, somewhat unexpectedly followed by Governments \& Politicians and Industry \& Emissions. We explain this phenomenon next by disentangling the labels by political leaning.

Figure~\ref{fig:exploratory:roles} reveals how the framing of a particular stakeholder category can vary with political leaning of the source.
Right-leaning media are more likely to depict Environmental Activists \& Organisations, and Legislation as the \vil and Industry and Emissions as either the \her or the \vic in the context of climate change news. Conversely, left-leaning media are more likely to frame Legislation as a \her, cover Environmental Activists less frequently overall, and predominantly frame the Industry as a \vil.

\section{Narrative Frame Prediction}
\label{sec:prediction}
Predicting narrative frames automatically and with high quality would open new possibilities for scaling media framing analyses to larger data sets, longer time spans or more languages. Given the political sensitivity of automated media analysis, models should not only be reliable but also transparent in their predictions. To this end, we present Retrieval-Based Frame prediction (RBF), which incorporates an embedding-based retrieval module into supervised classifiers. We compare RBF against a range of neural classifiers on multi-label frame prediction. RBF not only outperforms off-the-shelf fine-tuned transformers on this task, but also increases interpretability by predicting frames for a given article together with the most relevant article sentences for the frame as evidence. 
Section~\ref{sec:discussion} discusses additional modelling tasks supported by our data set to be addressed in future work.

\subsection{RBF: Retrieval-Based Frame Prediction}
We propose a simple method, retrieval-based frame prediction (RBF), which combines pre-trained language model embeddings with a retrieval objective. Similar approaches have been previously proposed in the context of word-sense disambigutaion and semantic frame predictions~\cite{jiang-riloff-2021-exploiting,blevins-zettlemoyer-2020-moving}. We embed (i)~short frame descriptions~$f_1\dots f_C$\footnote{RE: Solution or alleviation of the problem, CO: Human interest, emotion or dramatization of events, HI: Conflict or disagreement between two or more sides, MO: Morality or religion, EC: Economic consequences} and (ii)~sentences from an input news article $s_1 \dots s_N$ in a joint space, and retrieve sentences most proximate to the frame embedding:
\begin{equation}
\begin{aligned}
  h_i^s &= emb(s_i)\\
  h_j^f &= emb(f_i)\\
  rel(s_i, f_j) &= cos(h_i^s, h_j^f),
\end{aligned}
\end{equation}
where $h_i^s$ and $h_j^f$ are the embeddings of sentence $s_i$ and frame $f_j$, respectively, the relevance $rel$ of $s_i$ to $f_j$ corresponds to their cosine similarity. We use  SentenceBert~\cite{reimers2019sentence} as our embedding method $emb$.  Given an article, we obtain $J$ frame-specific relevance-rankings of all sentences in the input article.

We then train a linear classifier to predict the presence or absence of a frame in an article based on the most relevant sentences by our measure above. We include five input channels: channels (1)--(3)~are the three sentences most relevant for a frame according to RBF relevance; channel (4)~includes all sentences exceeding relevance threshold $\theta>0.15$, except for sentences (1)--(3), concatenated with a [SEP] token;\footnote{The threshold $\theta$ was tuned on the dev set in preliminary experiments.} and (5) contains the news article truncated at 256 tokens. Each channel is encoded with the Longformer~\cite{beltagy2020longformer}, final hidden state embeddings are concatenated and passed into the classifier. Longformer parameters are fine-tuned during the training process. We evaluate the importance of different channels by ablating the impact of the full article channel (5) (RBF -a) and additionally the threshold sentence channel (4) \mbox{(RBF -a,t)}. 

RBF combines two desiderata: First, it identifies multiple sentences relevant to a target frame, capturing key evidence that may be distributed across the article rather than locally. Second, RBF's frame-based sentence retrieval backbone can be interpreted as an `explicit attention mechanism', customized to the frame label to be predicted, with sentences serving as evidence. We evaluate the retrieved sentences in terms of their interpretability in Section~\ref{ssec:qual}.




\begin{table}[t]
    \centering
\begin{small}
\begin{tabular}{lcccc}
 \toprule
 Model        & Macro-Pr & Macro-Re & F1 \\\midrule
 Random       & 0.33 ($\pm$0.025)&0.49 ($\pm$0.023)& 0.39 \\
 Majority     & 0.14 ($\pm$0.030)&0.24 ($\pm$0.080)& 0.18 \\\midrule
 KNN          & 0.40 ($\pm$0.049)&0.61 ($\pm$0.071)& 0.49 \\
 BERT         & 0.44 ($\pm$0.109)&0.63  ($\pm$0.059)& 0.52 \\
 Longformer   & 0.48 ($\pm$0.054)& 0.63  ($\pm$0.052)&  0.53 \\
 Snippext     & {\bf0.56} ($\pm$0.074)&0.65 ($\pm$0.074)& 0.60 \\
 RBF        & {0.51 ($\pm$0.044)} &{\bf 0.76} ($\pm$0.131) & {\bf 0.61} \\\midrule
 RBF -a     & 0.50 ($\pm$0.047) & 0.70 ($\pm$0.077) & 0.58 \\
 RBF -a,t   & 0.48 ($\pm$0.041) & 0.63 ($\pm$0.085) & 0.55 \\\bottomrule
 \end{tabular}
\end{small}
    \caption{Frame prediction results of baselines (top) representative supervised and semi-supervised methods, and RBF (middle), as well as an ablation of RBF channels (bottom). We report macro-averaged precision and recall across the five labels with standard deviation (in brackets), and their harmonic mean (F1).}
    \label{tab:prediction_results}
\end{table}

\subsection{Experimental Setup}
Given the small size of the Narrative Frames Corpus, we adopt the simplest formalization of multi-label classification: for each model class (row in Table~\ref{tab:prediction_results}) we train five  individual binary classifiers, one per frame label.\footnote{This outperformed multi-label classifiers which shared a subset of parameters in preliminary experiments, presumably because the small number of article prevented models from learning useful interactions.}

\paragraph{Data set} 
We split the Narrative Frames Corpus set into five random folds of 60/20/20 train/dev/test data. As the proportion of articles mentioning each frame varies widely (see Appendix~\ref{app:labelstats}), within each fold, we balanced the number of articles that do/do not contain each frame. E.g., for a frame that was featured in the majority of articles, we randomly up-sampled articles {\it not} featuring the frame to a ratio of~1:1.

\paragraph{Comparison models}
We compare our method against 1. a {\bf Random} baseline; 2. a {\bf Majority} baseline which predicts the majority class per frame (1 for frames which occur in the majority of labelled articles, 0 for others); 3. a non-neural method which embeds articles based on TF-IDF representations, and trains one {\bf KNN} classifier per frame; 4. {\bf BERT-medium}~\cite{devlin2019bert} fine-tuned for binary frame prediction; 5. {\bf Longformer-base-4096}~\cite{beltagy2020longformer} fine-tuned for binary frame prediction; and 6. an adaptation of the {\bf Snippext} model~\cite{miao2020snippext,berthelot2019mixmatch}, a method for semi-supervised fine-tuning of pre-trained language models which was originally proposed in computer vision, but recently adapted to semi-supervised opinion mining~\cite{miao2020snippext}. Snippext fine-tunes BERT using an interpolation of a small amount of gold-labeled data, and a much larger set of unlabeled data with predicted, soft labels,\footnote{The original MixMatch additionally applies data augmentation to enhance consistency, which we disable due to the difficulty of DA in language, and for model simplicity.} drawing on the MixMatch strategy recently proposed in computer vision~\cite{berthelot2019mixmatch}. We augment our small labeled training data set with the $\approx$17.5K unlabelled climate-related articles (cf.,~Section~\ref{sec:annotation}). The input to all transformer models is truncated to 256 tokens.\footnote{In preliminary experiments we tested truncation at \{125, 256, 512 and 1024\}. We found that 256 worked best, presumably reflecting the pyramid structure of news with important information presented upfront/in the lead paragraphs. Very few articles were longer than 1024 tokens.} Detailed training settings and model parameters are provided in Appendix~\ref{app:parameters}.

\paragraph{Metrics} We evaluate models on correctly predicting the {\it presence} of frames in articles. We report macro-averaged precision and recall over the five frames, assigning equal importance to each frame label; as well as their harmonic mean (F1 score). 

\begin{table*}[ht]
\begin{small}
\begin{tabular}{lp{0.91\textwidth}}
\toprule
RE & (1) However the study finds that no single solution will avert the dangers, so {\bf a combined approach is needed}. 
(2) The key element is that these three {\bf solutions must be implemented together.}" 
(3) We also looked at increasing {\bf the efficiency of water use}, and we looked at {\bf better monitoring and recycling of fertiliser} - lots of it is lost and it runs off into rivers and causes dead zones in the oceans." 
\\\midrule
CO &  (1) {\bf Their answers – and reactions to them} – foreshadowed the {\bf fight} ahead with conservatives and industry regardless of who becomes the next president. 
(2) Democrats vying for president {\bf revealed a fundamental split over} how {\bf aggressively} the US should tackle climate change [\dots] in a seven-hour town hall meeting on Wednesday.  
(3) [\dots] held after the Democratic National Committee {\bf refused to sanction an official climate debate between candidates } and amid unprecedented {\bf pressure from young activists and the Democratic voting base} to tackle the climate crisis. 
 \\\midrule
HI & (1) {\bf To Janet}, this is a moral issue. 
(2) These were matters that we have historically agreed on, if for no other reason than {\bf the sake of our children and grandchildren's future.} 
(3) And {\bf in Jordan's case}, that would be Social Security, Medicare, education, health care and the like: Programs that benefit folks in his district. 
\\\midrule
MO & (1) It is, after all, the {\bf measure of one's moral fitness to value some things} (say, forgiveness) over others (vengeance). 
(2) {\bf When organized religion fades} and its would-be adherents are left to search for meaning, does the {\bf god of the environment} end their {\bf search for a moral authority}?
(3) That statement not only describes {\bf Judas's moral disorder} but also reminds the audience that any concern, {\bf holy as it may be} — poverty reduction, environmental protection, or any other {|\bf earthly mission} — that does not give a preferential deference to {\bf God, His creation, and acts of beauty such as that of Mary Magdalene} are sure signs of misaligned priorities. 
\\\midrule
EC &  (1) Both can impact the relative {\bf financial attractiveness of future energy options.}
(2) "Even Milton Friedman understood the existence of market externalities, the fact that damage to our environment is {\bf not accounted for in the free market} without {\bf placing some sort of price signal}." 
(3) They end up {\bf helping certain wealthy people to the disadvantage of the less fortunate.}" \\\bottomrule
\end{tabular}
\end{small}
\caption{Three top relevant sentences (right) extracted by RBF for articles which were correctly predicted as containing the frame (left), in order of decreasing relevance. Highlights of relevant phrases manually added in bold.}
\label{tab:example_predictions}
\end{table*}

\subsection{Main results}
\label{ssec:results}
Table~\ref{tab:prediction_results} shows the frame prediction results. All models significantly outperform the random and majority baselines. All neural methods perform better than the non-neural KNN. BERT performs worse than the Longformer, presumably due to the Longformer's higher capacity with 1.5$\times$ the parameters of BERT. RBF is best overall, suggesting that combining Longformer embeddings with a relevance based sentence retrieval backbone helps the models to focus on frame-relevant context.

We ablate the impact of the different channels in RBF in Table~\ref{tab:prediction_results} (bottom). The model performance drops with the removal of each input channels, suggesting that the input channels are complementary and each contributes to the performance. Snippext and RBF perform comparably, with inverse emphasis on precision and recall, however, only RBF offers explicit evidence for prediction (which we explore in the next section). A semi-supervised extension of RBF is a promising avenue for future work.

Given the multi-label nature of our data set, a natural question is how often models predict all and only the annotated frames for an article (exact match). RBF does so 18\% of the time.  Appendix~\ref{app:detailed_results} provides more detailed results and analyses of per-frame and per-label performance.

\subsubsection{Qualitative analysis}
\label{ssec:qual}
For each frame, we inspect sentences retrieved as highly relevant by RBF. Table~\ref{tab:example_predictions} displays these sentences. We boldfaced the most relevant phrases for ease of exposition. The selected sentences align closely with the definition of each frame: for \hi, they refer to the struggle of affected individuals and evoke empathy; for \mo they refer to god, religion and moral values; and for \ar they mention explicit solutions. One intriguing direction for future work will be to study the differences in {\it manifestation} of different frames across outlets from different sides along the political spectrum.

\section{Discussion}
\label{sec:discussion}
What are the recurring narratives that frame the public discourse about contested issues like climate change? Existing NLP approaches to frame prediction fall short of answering this question due to a focus on localized signals. Drawing on theories from the social and communicative sciences, we introduced a set of narrative framing devices to NLP, and integrated them with narrative roles assigned to the central entities in the news articles.

We applied our framework to the issue of climate change, and annotated ${>}$400 English-language news articles from major outlets with different political leanings with multi-label frames and narrative roles of entities and their stakeholder categories. 
Our exploratory analysis demonstrated how our framework can be utilized to study multiple levels of framing, including differences across outlets; co-occurrence of frames and narrative roles; and assignment of narrative roles to stakeholder categories.

With the ultimate goal of scaling such analyses to larger, unlabeled data sets, we introduced RBF, an effective and interpretable retrieval-based frame classifier. The `explicit attention' module of RBF not only improved performance over its backbone, the vanilla Longformer, but also naturally provides evidence for its predictions as a list of relevance-ranked article sentences. 

Our work addresses a disconnect between the complexity of framing acknowledged in the communication science literature, and models of framing in NLP. As recently surveyed by~\citet{ali-etal-2022-survey}, NLP approaches to framing predominantly focus on topic models or frequency-based methods, leaning heavily on local lexical signals as indicators for the presence or absence of a single frame per unit of analysis. However, the framing of a news article typically emerges from indicators spread throughout the text; frames can co-exist and interact with each other within a single news story. This paper takes one step towards such an integrated notion of framing in NLP in considering narrative frames and roles at the article level and adopting multi-label task formalization. 

Our work and results suggest many avenues for future research. The Narrative Frames Corpus supports research on joint models of framing and entity roles: the presence of an entity with a specific role (e.g., the \her) should render the presence of certain frames (e.g., \ar) more likely. Conversely, frames like \co impact the probability of the existence of the number and kind of different roles (e.g., the \her and the \vil). A joint model of frames and narrative roles could incorporate role labels with soft confidence weights as latent signal into a frame classification model. 

Annotator (dis)agreement and aggregation of answers to indicator questions into frame tags provide fertile grounds for future work. We echo a line of recent work on acknowledging label variation as a signal of genuine complexity  rather than noise. This holds true particularly for complex tasks like frame annotation which inevitably retain a level of subjective variation~\cite{10.1162/tacl_a_00293,plank2022problem}. In this paper we aggregated indicator labels into a hard frame label by voting, however we release the raw annotations as part of our data set. Future work could explore soft aggregation methods, delineate genuine variation from noise, and adopt disagreement-aware models and evaluation metrics.

The comparatively small size of the Narrative Frames Corpus and the competitive semi-supervised Snippext suggest further exploration of semi-supervised approaches. Integrating RBF with a Snippext-inspired semi-supervised framework, most simply by soft labeling articles as a function of their retrieved sentences and RBF relevance scores, would allow to leverage large unlabeled data sets while retaining RBF's interpretability. Alternatively, one could adapt models from different domains, for instance by drawing on the literature of modeling narrative roles in folk tales~\cite{valls2014toward,jahan2021inducing}.

\section{Limitations}
We acknowledge a range of limitations of our work.

As discussed in Sections~\ref{sec:annotation} and \ref{sec:discussion}, overall annotator agreement ranged from fair (frame annotations) to low (entity role annotations). We do not view this as a limitation per se, again pointing to the recent literature on the value of human label variance pointing at a potential loss of valuable information if we overly focus on arriving at a single gold label per instance, with high confidence~\cite{plank2022problem,10.1162/tacl_a_00293}. Future modeling work involving entity labels should, however, carefully inspect the role label variation, and potentially remove or aggregate selected annotations, before incorporating the labels as signal into predictive models. We explicitly refrained from training model in this paper to avoid the risk of training a predictor on an unfavorable noise-to-signal ratio.

Our data set focuses on English-language news reports, sampled from 2017 to 2019 in mainstream media outlets in the US and UK, and as such focuses on cultures and communities which are already well-resourced and well studied. With climate change being a global challenge, broadening data sets, annotations and models to more languages is an important direction for future work. We explicitly caution against projecting annotations across languages without careful validation as we expect the manifestation of framing, views on entities (or sheer set of dominant entities) to vary widely across countries and communities.

Even within our English study, we acknowledge that the size of annotated data is small for NLP scales, and an extension in the future is desirable. A related current limitation is the focus on just a single issue (climate change) and validation of our narrative framing framework for other issues is an important direction for the future. Finally, the annotation process was slow and costly, relying on trained, highly educated annotators with constant monitoring, rendering larger scale annotations challenging, on the one hand. On the other hand, we will release upon acceptance our annotation procedure including the full codebook with instructions, which have been optimized over several rounds of annotations and we hope can support more efficient annotation in the future.

\section*{Ethics statement}
This study was approved by the University of Mel-
bourne ethics board (Human Ethics Committee
LNR 3B), Reference Number 2023-22109-37029-4, and data acquisition and analysis has been taken
out to the according ethical standards. We hired four local annotators who were paid an hourly rate of \$53 AU 
in line with the casual research assistant hourly rates set up in the University of Melbourne collective agreement. 

We will release the Narrative Framing Corpus comprising of 428 news articles annotated with frame labels, entities, their narrative roles and stakeholder categories. We also publish the raw (non-aggregated) annotations. Our data set builds on news articles from the NELA corpora 2017-2019, which were released to the public domain (license CC0 1.0).\footnote{2017: \url{https://doi.org/10.7910/DVN/ZCXSKG}; 2018:  \url{https://doi.org/10.7910/DVN/ULHLCB}; 2019: \url{https://doi.org/10.7910/DVN/O7FWPO}} We release our code and Narrative Frames Corpus under a MIT license.

\section*{Acknowledgements}
We gratefully acknowledge the work of our annotators Candy Chu, Aarushi Kaul, Dana Pjanic, and Lloyd Rouse and the helpful feedback of the anonymous reviewers.

\bibliography{tacl2021}
\bibliographystyle{acl_natbib}

\onecolumn

\appendix
\section{Article selection}
\label{app:keywords}
We identify climate-related news articles in the NELA corpora (2017--2019) by searching for keywords identified in the Wikipedia climate change glossary\footnote{https://en.wikipedia.org/wiki/Glossary\_of\_climate\_change}. A few generic terms were removed (`weather') as too broad for our query. We consider an article as relevant either if $>=$1 keywords are found in the article title or $>=$3 mentions of climate keywords are found in the article body.
    
\section{Codebook}
Our full codebook of 22 fine-grained questions presented to annotators. The questions used in the final factor analysis model and subsequent analyses are boldfaced. We show the indicator distribution, i.e., the fraction of annotated articles in which two or more annotators answered 'yes' to a given question, in brackets.


\label{app:codebook}
\begin{tabular}{p{0.6cm}p{5cm}p{9cm}}
\toprule
{\bf ID}  & {\bf Question}  & {\bf Annotation rules} \\\midrule
{\bf RE1} \newline (0.51)  & {\bf 2. Does the story suggest a solution(s) to the issue/problem?}   &{\bf  Mark ‘yes’ if (a) solution(s), or a strategy to mitigate the problem, is explicitly mentioned.}  \\
RE2  & 3. Is this problem/issue resolved in the story?   & Mark ‘yes’ if the story explicitly mentions that the problem has been resolved. \\
RE3  & 4. Is there any hope in the story for future resolution of the problem/issue?   & Mark ‘no’ if the story is about a failed attempt to tackle the issue under discussion. \\
RE4  & 5. Does the story suggest that the issue/problem requires urgent action?   & Mark ‘yes’ if there is an explicit call for action, or ongoing efforts to alleviate the problem are (1) explicitly described and (2) raise a sense of urgency  \\
{\bf RE5} \newline (0.55) & {\bf 6. Does the story suggest that some entity could alleviate the problem?    If your answer is "yes", please select the most appropriate entity. }  & {\bf Mark ‘yes’ if at least one entity in the story is described as actively alleviating, or planning to alleviate, the problem.  If multiple options:  select the one that’s most central/prevalent in the article (in terms of \#mentions / mentions in the central parts like title and opening.)}  \\
RE6  & 7. Does the story suggest that some entity is responsible for the issue/problem?   If your answer is "yes", please select the most responsible entity.   & Mark ‘yes’ if at least one entity in the story is described as actively causing, or having caused, the problem.  If multiple options:  select the one that’s most central/prevalent in the article (in terms of the number of mentions / mentions in the central parts like title and lead paragraphs).  \\\bottomrule
\end{tabular}

\begin{tabular}{p{0.6cm}p{5cm}p{9cm}}
\toprule
{\bf HI1} \newline (0.27) & {\bf 8. Does the story provide a human example or a "human face" on the problem/issue?}  & Select ‘yes’ if the story uses “dramatization” i.e., explicitly refers to how the issue impacts the personal life living entities (including animals). \\
{\bf HI2} \newline (0.21) & 9. {\bf Does the story employ adjectives or personal vignettes that generate feelings of outrage, empathy-caring, sympathy, or compassion? }    & Mark ‘yes’ if the story uses emotional language (original Q is already very concrete). \\
HI3  & 10. Does the story emphasize how one or more entities are NEGATIVELY affected by issue/problem?    If your answer is "yes", please select the most negatively affected entity.  & Select ‘yes’ if the story explicitly refers to how one or more entity/ies suffer from the problem/issue. \\
HI4  & 11. Does the story emphasize how one or more entities are POSITIVELY affected by the issue/problem?    If your answer is "yes", please select the most positively affected entity.  & Select ‘yes’ if the story explicitly refers to how one or more entity/ies benefit from the problem/issue.  \\
{\bf HI5} \newline (0.07) & {\bf 12. Does the story go into the private or personal lives of the entities involved?}  & Mark ‘yes’ if the story explicitly refers to the personal life of at least one entity. \\\bottomrule
\end{tabular}

\begin{tabular}{p{0.6cm}p{5cm}p{9cm}}
\toprule
{\bf CO1 } \newline (0.73) & {\bf 13. Does the story reflect disagreement between political parties/ individuals /groups/ countries? }    & Select ‘yes’ even if the story describes a disagreement or a conflict in a passive/observational manner. \\
{\bf CO2 } \newline (0.40) & {\bf 14. Does one party/ individual/ group/ country reproach another? }   & Select ‘yes’ if the story explicitly refers to the active conflict between two or more entities – past or present. \\
{\bf CO3} \newline (0.54)  & {\bf 15. Does the story refer to two sides or more than two sides of the problem or issue?}  & Select ‘yes’ if the story explicitly mentions at least two viewpoints on the current issue (even if they’re not presented in a balanced, objective manner). \\
CO4  & 16. Does the story refer to winners and losers?    If your answer is "yes", please select the most appropriate winner/loser entity.  & Select ‘yes’ if the story explicitly refers to one or more ‘winners’ and/or ‘losers’ which emerged from an active conflict/argument/war.  Note, in some stories an entity can be both a winner and a loser. \\\bottomrule
\end{tabular}

\begin{tabular}{p{0.6cm}p{5cm}p{9cm}}
\toprule
{\bf MO1} \newline (0.07) & {\bf 17. Does the story contain any moral message? }   & Select ‘yes’ if the story explicitly applies standards or judgments of right or wrong to entities, actions or events. \\
{\bf MO2} \newline (0.05)  & {\bf 18. Does the story make reference to morality, God, and other religious tenets?  }   & Select ‘yes’ if the story explicitly refers to religious tenets or moral obligations framed through the lens of obligations to a spiritual community. Select ‘yes’ also if the mention is indirect e.g., through a quote or a metaphor. \\
MO3  & 19. Does the story offer specific social prescriptions about how to behave?      & Select ‘yes’ if the story explicitly mentions expectations around norms of conduct, limitations or prohibitions on actions or events. \\\bottomrule
\end{tabular}

\begin{tabular}{p{0.6cm}p{5cm}p{9cm}}
\toprule
{\bf EC1} \newline (0.28) & {\bf 20. Is there a mention of financial losses or gains now or in the future? }       & Select ‘yes’ if the story explicitly refers to the financial impacts of the issue. \\
{\bf EC2} \newline (0.37) & {\bf 21. Is there a mention of the costs/degree of the expense involved? }                    & Select ‘yes’ if the story explicitly refers to the amount of loss or gain (e.g., “\$100,000”, “enormous cost”).  \\
{\bf EC3} \newline (0.25)& {\bf 22. Is there a reference to the economic consequences of pursuing or not pursuing a course of action?}   & Select ‘yes’ if the story explicitly mentions the impacts of action or inaction on the economy. \\\bottomrule
\end{tabular}

\section{Annotation Interface}
\label{app:interface}
\begin{figure}[th]
    \centering
    \includegraphics[width=\textwidth]{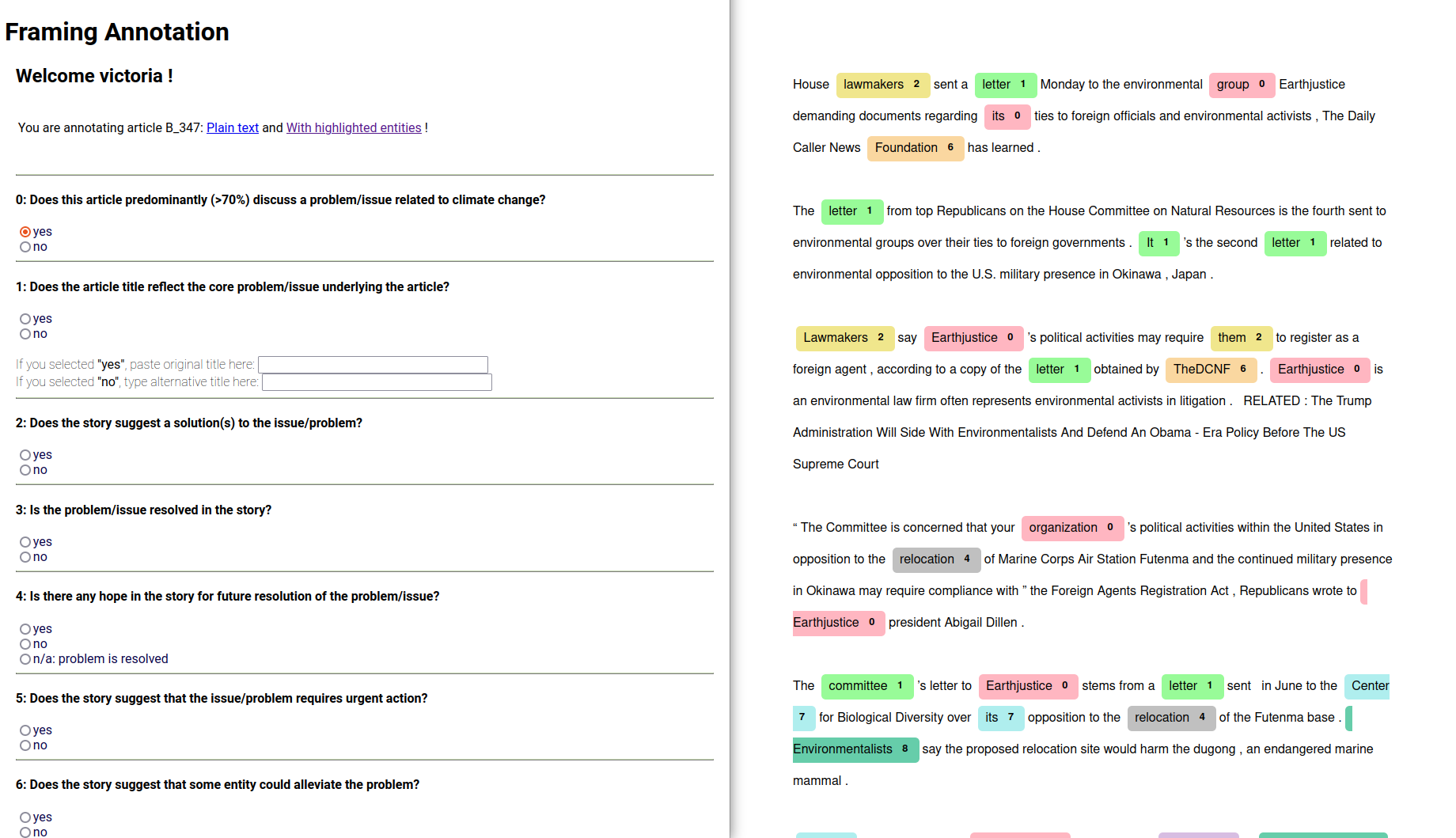}
    \caption{Our annotation interface. Left: Excerpt of the annotation form which covers the 22 binary indicator questions, and free-text fields to record  role-specific entities. Right: the news article with prevalent entities highlighted (based on automatic entity recognition and co-reference resolution.)}
    \label{fig:app:interface}
\end{figure}

Figure~\ref{fig:app:interface} shows our annotation interface split into an answer form (left) and an article display with (optional) highlighting of prevalent entities (right, in color).

\section{Media Outlets}
\label{app:outlets}
Table~\ref{app:tab:outlets} lists all media outlets in the labeled data set, together with number of articles and MBFC political leaning.

\begin{table}[h!]
    \centering
    \begin{tabular}{p{0.95\textwidth}}
    \toprule
{\bf left\_bias:} thehuffingtonpost (47), vox (22), cnn (18), politicususa (16), motherjones (10), shareblue (5), dailykos (4), talkingpointsmemo (3), slate (2), palmerreport (1)
\\\midrule
{\bf left\_center\_bias:} bbc (22), theguardian (20), npr (14), usatoday (10), cbsnews (9), yahoonews (9), thenewyorktimes (8), pbs (4), fusion (4), cnbc (3), thehill (3)\\\midrule
{\bf right\_bias:} thedailycaller (39), drudgereport (27), foxnews (13), theblaze (10), nationalreview (9), redstate (6), newsbusters (4), thepoliticalinsider (2), therightscoop (1)
\\\midrule
{\bf questionable\_source:} breitbart (38), rt (18), dailymail (12), bipartisanreport (2), theduran (1)
\\\bottomrule
    \end{tabular}
    \caption{Media outlets in our data, grouped by their political leaning (based on the Media Bias Fact Check portal; rows) and with associated article count (in brackets).}
    \label{app:tab:outlets}
\end{table}

\section{Entity groups}
\label{app:entitygroups}
Table~\ref{app:tab:entitygroups} lists our set of entity groups, together with some representative examples and the total number (tokens) of instances assigned to each group.

\begin{table}[h!]
    \centering
    \begin{tabular}{lcp{5cm}}
    \toprule
    {\bf Entity Group} & {\bf Count} & {\bf Example entities} \\\midrule
 GOVERNMENTS\_POLITICIANS\_POLIT.ORGS & 828  & democrats, trump, the EPA\\
 INDUSTRY\_EMISSIONS & 271  & fossil fuels, carbon pollution, plastic, capitalism\\
 LEGISLATION\_POLICIES\_RESPONSES & 251  & US climate response, Green New Deal, paris agreement\\
 GENERAL PUBLIC & 241  & americans, indigenous people, public health \\
 ANIMALS\_NATURE\_ENVIRONMENT & 213  & air quality, the ocean, the earth\\
 ENV.ORGS\_ACTIVISTS & 105  & Greta Thunberg, youth activists, extinction rebellion \\
 SCIENCE\_EXPERTS\_SCI.REPORTS & 59  & the stupid bloody academics, professors, University of Auckland study \\
 CLIMATE CHANGE & 52  & climate change, global warming\\
 OTHER & 49  & meat consuption, resilience, god \\
 AMBIGUOUS & 37  & deniers, response\\
 GREEN TECHNOLOGY\_INNOVATION & 31  &  wind energy, geoengineering, renewables\\
 MEDIA\_JOURNALISTS & 26  &  journalists, media outlets, CNN\\\bottomrule
    \end{tabular}
    \caption{Stakeholder groups with token count and examples of assigned entities.}
    \label{app:tab:entitygroups}
\end{table}

\section{Label statistics}
\label{app:labelstats}
\begin{figure}[h!]
    \centering
    \includegraphics[width=1\textwidth]{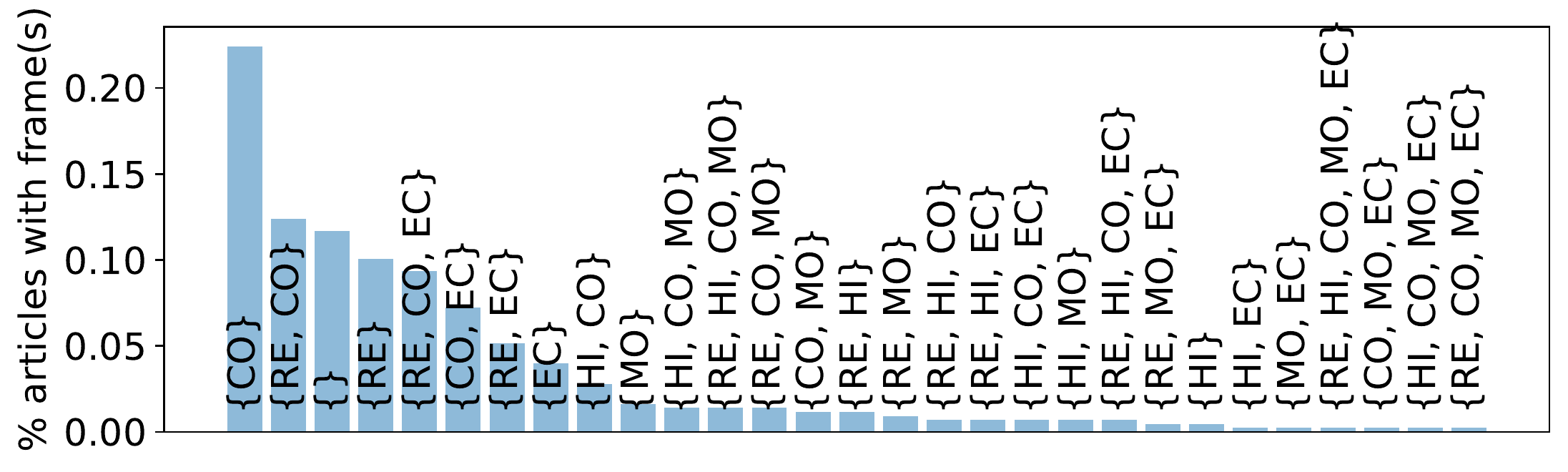}
    \caption{Distribution of the most frequent frame labels.}
    \label{fig:frameprev}
\end{figure}

Figure~\ref{fig:frameprev} shows label distributions in our data set. In terms of prevalence of our five frames individually, \co is present in 63\%; \ar in 45\%; \ec in 30\%; \hi in 11\% and \mo in 10\% of all annotated articles.

\section{Hyperparameters, model sizes and compute costs}
Table~\ref{tab:model_parameters} lists the hyperparameters for all neural models, and their total number of parameters.
\label{app:parameters}
\begin{table}[h!]
    \centering
    \begin{tabular}{c|ccccc}
    \toprule
    Item & RBF & BERT & Longformer & Snippext \\
    \midrule
    random\_seed & 1042 & 1042 & 1042 & -- \\
    learning\_rate  &  2e-6 & 2e-6 & 2e-6 & 2e-6 \\
    max\_len & 256 & 256 & 256 & 256 \\
    batch\_size & 8 & 16 & 16 & 8 \\
    epochs & 20 & 20 & 20 & 20 \\
    total \# parameters &149 M& 109 M& 149 M& 149M\\
    \bottomrule
    \end{tabular}
    \caption{Hyper parameters and model size for all neural models.}
    \label{tab:model_parameters}
\end{table}
For KNN, K was tuned on the dev set for each fold. L2 distance was used as a distance function and for all other parameters default values were used from this implementation: \url{https://scikit-learn.org/stable/modules/generated/sklearn.neighbors.KNeighborsClassifier.html}.\\
\\
All neural models were trained on a single Nvidia A100 GPU. The total compute budget covering all experiments were 26 hours.

\section{Additional Frame Prediction Results}
\label{app:detailed_results}

\paragraph{Frame-wise analysis} Table~\ref{tab:app:framelevel} shows frame-wise prediction performance averaged over five runs, for the three best performing models in Section~\ref{ssec:results}. We observe that \hi (HI) is the most challenging frame for all models, while access to the unlabelled data appears to be effective to boost performance for Snippext on this challenging class. The most prevalent \co (CO) frame is predicted most reliably. 

\begin{table*}[h!]
\begin{center}
\begin{tabular}{cccc|ccc|ccc}
\toprule
&\multicolumn{3}{c}{Longformer}&\multicolumn{3}{c}{RBF}&\multicolumn{3}{c}{Snippext}\\
&Precision&Recall&F1&Precision&Recall&F1&Precision&Recall&F1\\\midrule
RE&0.52&0.80&0.63&0.52&0.83&0.64&0.54&0.71&0.61\\
HI&0.22&0.28&0.24&0.20&0.59&0.30&0.48&0.47&0.48\\
CO&0.70&0.92&0.80&0.70&0.89&0.78&0.76&0.83&0.80\\
MO&0.50&0.46&0.48&0.51&0.69&0.59&0.51&0.55&0.53\\
EC&0.46&0.70&0.55&0.63&0.80&0.70&0.51&0.68&0.58\\\bottomrule
\end{tabular}
\end{center}
\caption{Frame-level prediction results of the three best performing models in Section~\ref{ssec:results}, averaged over five folds.}
\label{tab:app:framelevel}
\end{table*}

\paragraph{Label-wise performance} We analyze for RBF, which multi-labels are predicted best (worst). Overall, RBF achieves 18\% exact match accuracy. Table~\ref{tab:app:labellevel} lists the best (left) and worst (right) performing multi-labels, with an occurrence (N) of at least three in the gold labeled data set. 
\begin{table}[h!]
\centering
\begin{tabular}{lrlccclrl}
\toprule
\multicolumn{3}{c}{Best}&&&&\multicolumn{3}{c}{Worst}\\
Label & N & \% correct&&&&Label & N & \% correct\\\midrule
RE;CO;EC&40&0.4&&&&HI;CO&12&0.0\\
RE;CO&53&0.34&&&&MO&7&0.0\\
RE;HI;CO;EC&3&0.33&&&&RE;CO;MO&6&0.0\\
RE&43&0.21&&&&RE;HI;CO;MO&6&0.0\\\bottomrule
\end{tabular}
\caption{The five best-predicted multi-labels by RBF (left) and worst-predicted (right) with ${>}1$ occurrences in the labeled data. For each label we show its prevalence in the data set (N) and \% predicted exact match.}
\label{tab:app:labellevel}
\end{table}
We can see, in line with Table~\ref{tab:app:framelevel} that the model struggles with multi-labels involving the \hi or \mo frame. We also find that performance does not correlate with the number of frames: both the top and bottom performing sets include single-labels and high multi-labels.

\subsection{Error analysis}
\label{app:erroranalysis}
\paragraph{Overprediction of \hi}
We first analyze the false positive prediction of \hi (HI). We list sentences extracted by RBF most confidently extracted as evidence for an incorrect positive HI label, some examples include context for clarity, in \textcolor{gray}{gray}:
\begin{center}
\begin{tabular}{lp{0.9\textwidth}}
(i) &``{\it In its research, SPARK Neuro measured physiological data such as brain activity and palm sweat to quantify people's emotional reactions to stimuli.}'' (Breitbart, 2019-09-02)\\
(ii) &``\textcolor{gray}{`Something must have changed in the debate that so many young people are speaking up and so many young people are being targeted,' Thunberg told Yahoo News in response to the mockery her movement has received from leaders like Trump and Russian President Vladimir Putin.} {\it `They can sense that we are making an impact'} '' (Yahoo News, 2019-10-17)\\
(iii) &``\textcolor{gray}{[...] it might be tempting to criticize or dismiss activists supporting it. But Amy Myers Jaffe hopes older, more experienced policymakers won't do that. `We need not to discourage them,' she says.} {\it They have an energy and will to innovation that is not only infectious, but inspiring.}'' (NPR, 2019-02-08)\footnote{Grammatical error is present in the original article}
\end{tabular}
\end{center}
Sentence (i) refers to emotions, which are often linked to a human interest perspective, but not in this quote which refers to a scientific study. In case (ii) `they' refers to members of the Russian government, pointing to an error in  contextualization. Resolving `they' as a cataphora to `so many young people' may be plausible in a superficial, yet incorrect read, as evidence for a \hi frame.
Sentence (iii) is about shaping a strategic response to activists, and does not imply any human impact or emotion.

\paragraph{Overprediction of \ar} is often due to quotes, or criticisms, of previously proposed policies without framing them as actual solutions in the context of the article, for instance
\begin{center}
\begin{tabular}{lp{0.9\textwidth}}
(iv)& ``{\it In the 2016 presidential election, Sanders staked out the most ambitious climate platform of any candidate, 
vowing to slash carbon dioxide pollution 40 percent by 2030, end fossil fuel subsidies and ban fracking. }'' (Huffington Post, 2018-04-12)
\end{tabular}
\end{center}
where Sanders' ambitions are discussed in the context election campaigns of different candidates.

\paragraph{Underprediction of \mo}. Moral issues can take very different forms, for instance an article from The Daily Caller (2018-12-12) refers to the importance of `{\it tolerance of [a] diversity of viewpoints}', and automatic prediction of a link to moral questions is near impossible without a fundamental understanding of the concept of moralilty. Millenia of theoretical arguments on the conceptualization of moral, and a much shorter yet active stream of research in NLP/ML confirms the intrinsic challenge with this frame~\cite{xie-etal-2019-text,ziems-etal-2022-moral}. Our model to a large extent relies on explicit mentions of the keyword `moral' or religious terms.

\end{document}